\documentclass[manuscript,screen]{acmart}
\AtBeginDocument{%
  \providecommand\BibTeX{{%
    \normalfont B\kern-0.5em{\scshape i\kern-0.25em b}\kern-0.8em\TeX}}}

\setcopyright{acmcopyright}
\copyrightyear{2018}
\acmYear{2018}
\acmDOI{10.1145/1122445.1122456}

\acmConference[Woodstock '18]{Woodstock '18: ACM Symposium on Neural
  Gaze Detection}{June 03--05, 2018}{Woodstock, NY}
\acmBooktitle{Woodstock '18: ACM Symposium on Neural Gaze Detection,
  June 03--05, 2018, Woodstock, NY}
\acmPrice{15.00}
\acmISBN{978-1-4503-XXXX-X/18/06}

\begin{document}

\title{xCos: An Explainable Cosine Metric for Face Verification Task}

\author{Yu-Sheng Lin}
\email{biolin@cmlab.csie.ntu.edu.tw}
\affiliation{%
  \institution{National Taiwan University}
  \city{Taipei}
  \country{Taiwan}
}

\author{Zhe-Yu Liu}
\affiliation{%
  \institution{National Taiwan University}
  \city{Taipei}
  \country{Taiwan}}
\email{zhe2325138@cmlab.csie.ntu.edu.tw}

\author{Yu-An Chen}
\affiliation{%
  \institution{National Taiwan University}
  \city{Taipei}
  \country{Taiwan}
}
\email{r07922076@cmlab.csie.ntu.edu.tw}

\author{Yu-Siang Wang}
\affiliation{%
 \institution{University of Toronto}
 \city{Toronto}
 \country{Canada}}
\email{yswang@cs.toronto.edu}

\author{Ya-Liang Chang}
\affiliation{%
 \institution{National Taiwan University}
 \streetaddress{Rono-Hills}
 \city{Taipei}
 \country{Taiwan}}
\email{yaliangchang@cmlab.csie.ntu.edu.tw}

\author{Winston H. Hsu}
\affiliation{%
  \institution{National Taiwan University}
  \city{Taipei}
  \country{Taiwan}}
\email{whsu@ntu.edu.tw}

\renewcommand{\shortauthors}{Lin, et al.}

\begin{abstract}
  We study the XAI (explainable AI) on the face recognition task, particularly the face verification here. Face verification is a crucial task in recent days and it has been deployed to plenty of applications, such as access control, surveillance, and automatic personal log-on for mobile devices. With the increasing amount of data, deep convolutional neural networks can achieve very high accuracy for the face verification task. Beyond exceptional performances, deep face verification models need more interpretability so that we can trust the results they generate. In this paper, we propose a novel similarity metric, called explainable cosine ($xCos$), that comes with a learnable module that can be plugged into most of the verification models to provide meaningful explanations. With the help of $xCos$, we can see which parts of the two input faces are similar, where the model pays its attention to, and how the local similarities are weighted to form the output $xCos$ score. We demonstrate the effectiveness of our proposed method on LFW and various competitive benchmarks, resulting in not only providing novel and desiring model interpretability for face verification but also ensuring the accuracy as plugging into existing face recognition models.
\end{abstract}




\setcopyright{acmcopyright}
\acmJournal{TOMM}
\acmYear{2021} \acmVolume{1} \acmNumber{1} \acmArticle{1} \acmMonth{1} \acmPrice{15.00}\acmDOI{10.1145/3469288}

\maketitle

\section{Introduction}
%
%
%
%
Recent years have witnessed rapid development in the area of deep learning and it has been applied to many computer vision tasks, such as image classification \cite{imagenet,resnet}, object detection \cite{fasterrcnn}, semantic segmentation \cite{fcn}, and face verification \cite{SunCWT14}, \textit{etc}. In spite of the astonishing success of convolutional neural networks (CNNs), computer vision communities still lack an effective method to understand the working mechanism of deep learning models due to their inborn non-linear structures and complicated decision-making process (so-called ``black box''). Moreover, when it comes to security applications (e.g., face verification for mobile screen lock), the false-positive results for unknown reasons by deep learning models could lead to serious security and privacy issues. The aforementioned problems will make users insecure about deep learning based systems and also make developers hard to improve them. Therefore, it is crucial to increase transparency during the decision-making process for deep learning models.  
A rising field to address this issue is called explainable AI (XAI) \cite{gunning2017explainable}, which attempts to empower the researcher to understand the decision-making process of neural nets via explainable features or decision processes. With the support of explainable AI, we can understand and trust the neural networks' prediction more. In this work, we focus on building a more explainable face verification framework with our proposed novel $xCos$ module. With $xCos$, we can exactly know how the model determines the similarity score via examining the local similarity map and the attention map.

\begin{figure*}
  \centering
  \includegraphics[width=0.87\linewidth]{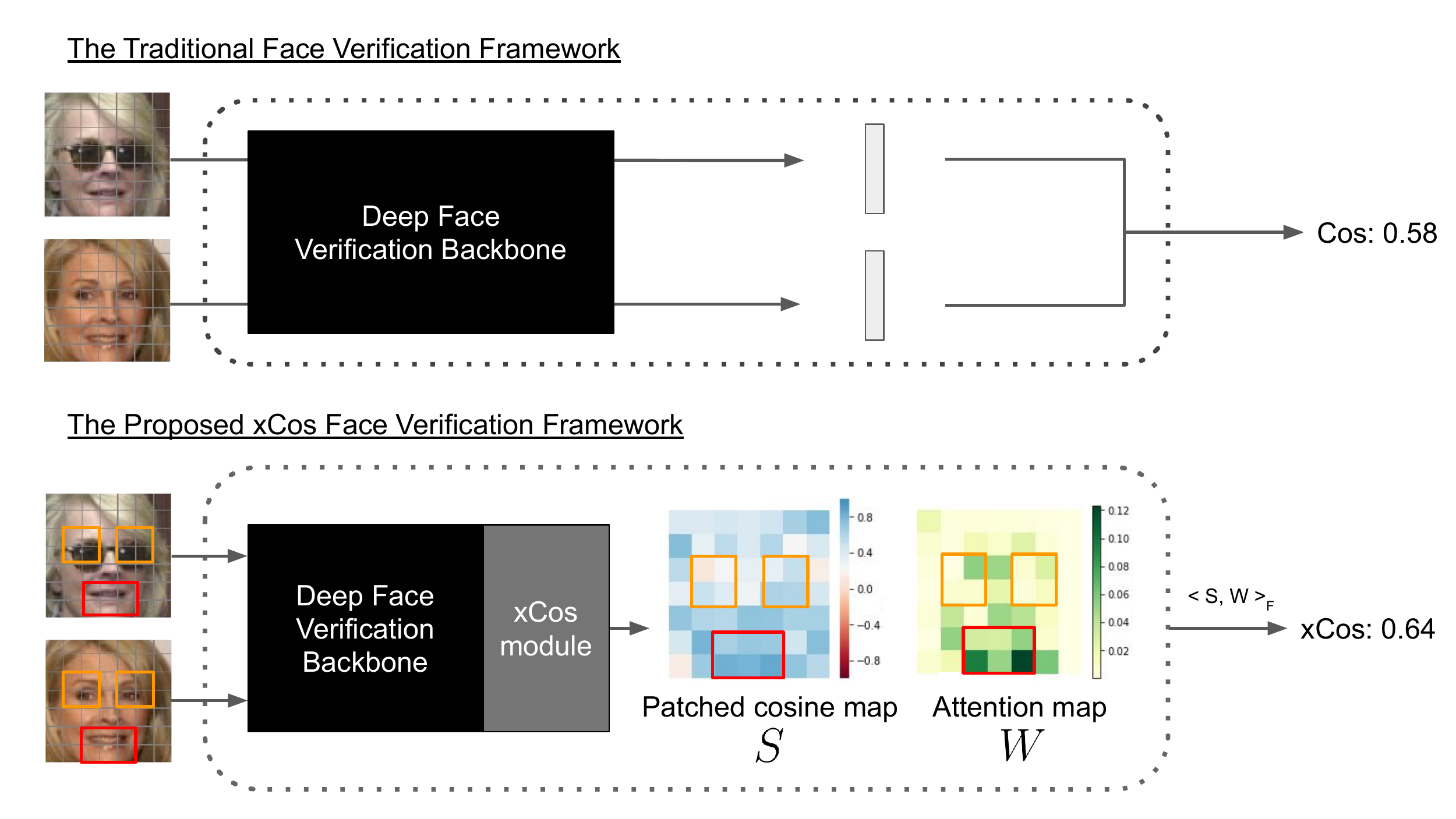}
  \caption{\textbf{Example of $xCos$ framework.} 
  Traditional face verification models provide no spatial clues about why the two images are the same identity or not.
  The models equipped with our proposed $xCos$ module allow the user to visualize the similarity map between two people for each part of a face and our model cares to produce the final similarity score, $xCos$ (explainable cosine). The $<S, W>_{F}$ denotes the Frobenius inner product between $S$ and $W$. We can see that $xCos$ module can be plugged into any existed deep face verification models and the existed face verification models can be more easily interpreted with our proposed $xCos$.
  }
  \label{fig:proposedIdea}
\end{figure*}

We begin our work with a pivotal question: ”How can the model produce more explainable results?” To answer this question, we first investigate the pipeline of current face verification models and then introduce the intuition of the human decision-making process for face verification.

Next, we formulate our definition of interpretability and design the explainable framework that meets our needs. 

State-of-the-art face verification models \cite{deng2018arcface,SphereFace} extract deep features of a pair of face images and compute the cosine similarity or the L2-distance of the paired features. Two images are said to be from the same person if the similarity is larger than a threshold value. However, with this standard procedure, we can hardly interpret these high dimensional features with our knowledge. Although some previous works \cite{Grad-CAM,Grad-CAM++,FG-18_Castan2018VisualizingAQ} attempt to visualize the most salient features, the saliency maps produced by these methods are mostly used to locate objects in a single image rather than interpret the similarity of two faces. In contrast, our framework interprets the verification result by combining the local similarity map and the attention map. (cf. Fig.~\ref{Fig:grad-CAM}) With the proposed method, we can strike a balance between verification accuracy and visual interpretability.  

We observe that humans usually decide whether the two face images are from the same identity by comparing their face characteristics. For instance, if two face images are from the same person, then the same parts of the two face images should be similar, including the eyes, the nose, etc. Based on this insight, we develop a novel face verification framework, $xCos$, which behaves closely to our observation. 

Illustrated by the observation above, we define the \textbf{interpretability} in the face verification that the output similarity metric aims to provide not only the local similarity information but also the spatial attention of the model. Based on our definition of interpretability, we propose a similarity metric, $xCos$, that can be analyzed in an explainable way. As shown in Fig.~\ref{fig:proposedIdea}, we can insert our novel $xCos$ module\footnote{The module is publicly available at https://github.com/ntubiolin/xcos} into any deep face verification networks and get two spatial-interpretable maps. Here we plug the proposed $xCos$ module into ArcFace \cite{deng2018arcface} and CosFace \cite{Wang2018CosFaceLM}. The first map displays the cosine similarity of each grid feature pair, and the second one shows what the model pays attention to. With the two visualized maps, we can directly understand which grid feature pair is more similar and important for the decision-making process.  

The main contributions of this work are as follows:

\begin{itemize}
\item We address the interpretability issue in the face verification task from the perspective of local similarity and model attention, and propose a novel explainable metric, \textbf{$xCos$ (explainable cosine)}.
\item We treat the convolution feature as the face representation, which preserves location information while remaining good verification performances.
\item The proposed $xCos$ module can be plugged into various face verification models, such as ArcFace \cite{deng2018arcface} and CosFace \cite{Wang2018CosFaceLM} (cf. Table~\ref{tab:LFW_accuracy}).

\end{itemize}
\section{Related Work}
\subsection{Face Verification}
The face verification task has come a long way these years. GaussianFace \cite{GaussianFace} first proposed Discriminative Gaussian Process Latent Variable Model that surpasses human-level face verification accuracy. Due to the emerging of deep learning, DeepFace \cite{Parkhi2015DeepFR}, SphereFace \cite{SphereFace}, CosFace \cite{Wang2018CosFaceLM}, and ArcFace \cite{deng2018arcface} achieve great performances on the face verification task with different loss function designs and deeper backbone architectures. However, there are still challenging scenarios that might cause a verification failure like cross-age \cite{CACD, calfw} or occlusion \cite{ARdatabase}. 
In \cite{Age_invariant_face_recognition}, the face images of different ages are treated as a face time series, and then the Multi-Features Fusion and Decomposition (MFFD) model is applied to solve the Age-Invariant Face Recognition task.
 Faced with the incorrect verification result, the user can hardly understand the cause of the failure.
\begin{figure*}[!ht]
  \centering
  \includegraphics[width=1.0\linewidth]{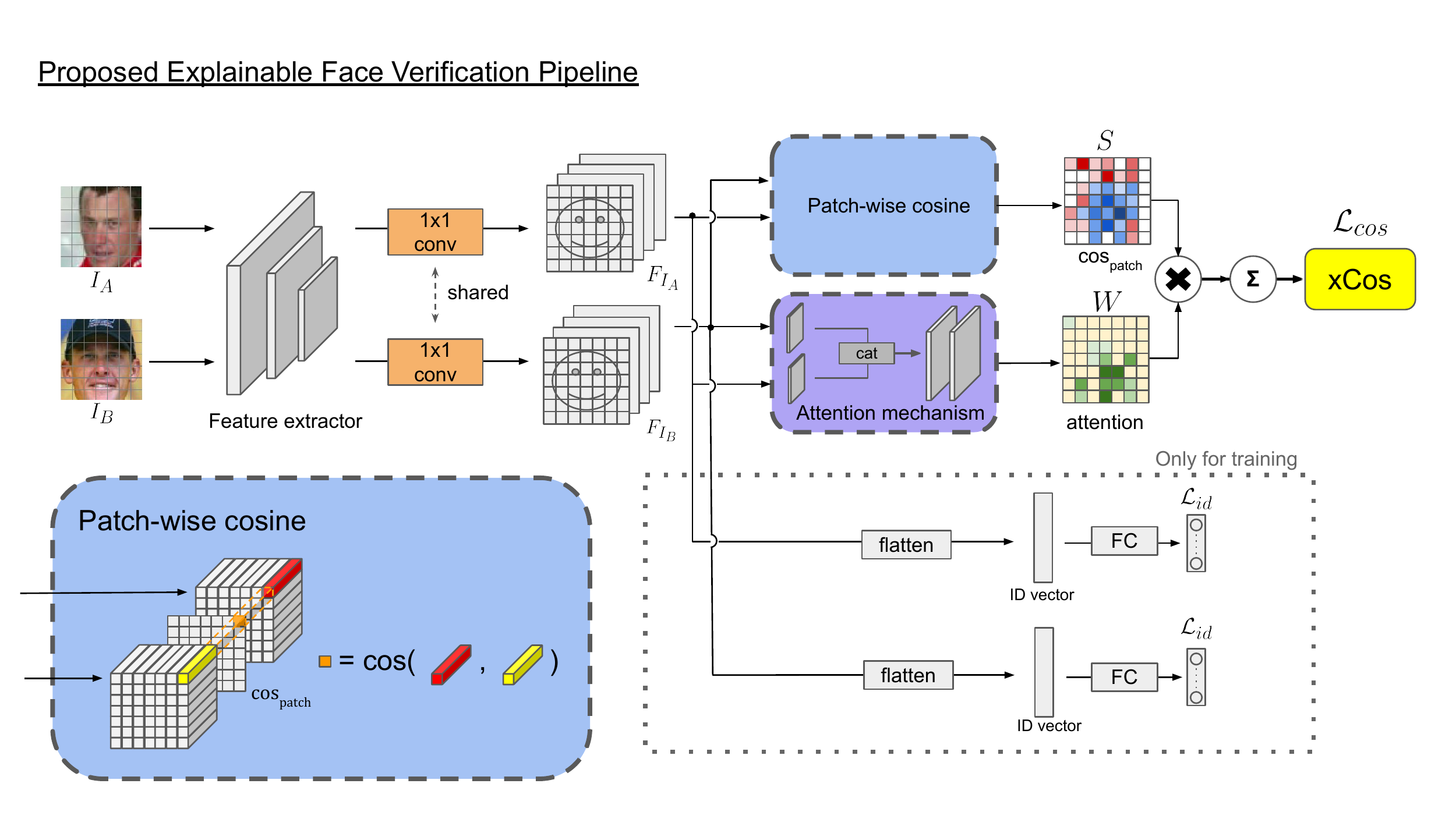}
  \caption{\textbf{Proposed Architecture.}
  Our proposed architecture contains one modified CNN backbone and two branches for $xCos$ and identification. The CNN backbone is responsible for extracting face feature for each identity. To preserve the position information of each feature point, the final flatten and fully-connected layers of the backbone (e.g., ArcFace \cite{deng2018arcface} or CosFace \cite{Wang2018CosFaceLM}) are replaced with an 1 by 1 convolution.
  On the $xCos$ branch, we compute one patched cosine map $\textbf{S}$ (i.e. $cos_{patch}$ in the figure) by measuring the cosine similarity element-wisely between the two feature maps of compared images. Meanwhile, an attention weight map $\textbf{W}$ is generated by our attention mechanism based on the two feature maps. The patched cosine map $\textbf{S}$ is then weighted summed according to the attention weight map $\textbf{W}$ to get the final $xCos$ similarity value. The $xCos$ is supervised under the cosine similarity generated by another face recognition model like ArcFace.
  The identification branch flattens the extracted feature and passes it into another fully connected layer for ID prediction. The loss $L_{id}$ is used to stabilize the training process and can be any common face recognition loss like the one in ArcFace.}
  \label{fig:architecture}
\end{figure*}
\subsection{Explainable AI}
With the rising demand for explainable AI, there have been plenty of works related to this topic in recent years.  Visualizations of convolution neural networks using saliency maps are the main techniques used in \cite{LearningDeepFeaturesforDiscriminativeLocalization,Grad-CAM,Grad-CAM++}. In \cite{VINET}, the importance estimation network produces a saliency map for every prediction so that doctors can make accurate diagnoses with the diagnostic visual interpretation. However, as we have mentioned, the saliency map is more suitable for locating objects. Knowledge Distillation \cite{DistillingTheKnowledgeInANeuralNetwork} is another path to interpretable machine learning because we can transfer the learned knowledge from the teacher model to the student model. \cite{Liu2018ImproveDNNwithKnowledgeDistilling} realizes this idea through distilling Deep Neural Networks into decision trees. In our work, the current face verification model functions as the teacher model to supervise the $xCos$ module with the cosine similarity values it produces.

Decomposing the deep feature into interpretable components can reveal how the model makes decisions.
BagNet \cite{brendel2018bagnets} combines the bag-of-local-feature concept with convolution neural network models and performs well on ImageNet. By classifying images based on the occurrences of patched local features without considering their spatial ordering, Bagnet \cite{brendel2018bagnets} provides a straightforward way to quantitatively analyze how exactly each patch of the image impacts the classification results. 

In \cite{Explain_semantically_and_quantitatively_Chen_2019_ICCV}, the proposed method generates a report that quantitatively describes which visual semantic parts contribute the most. Although \cite{Explain_semantically_and_quantitatively_Chen_2019_ICCV} also explains the CNN model by decomposing the output into different visual concepts, it might be not easy to apply \cite{Explain_semantically_and_quantitatively_Chen_2019_ICCV} on modern face verification models due to the difficulty in generating task-specific visual concepts. In comparison, the visual concept of our proposed method is defined as how similar two face parts are, which needs neither annotation labor nor a pretrained concept extractor. Therefore, xCos might be more suitable for the face verification task.

In \cite{ReID_Yang_2019_CVPR}, the class activation maps augmentation is used to discover discriminative visual cues by applying overlapped activation penalty. The difference between our proposed method and \cite{ReID_Yang_2019_CVPR} is the information in the heatmap. The heatmap in \cite{ReID_Yang_2019_CVPR} indicates the feature similarity, while the heatmap in our proposed method represents the importance of each local similarity. The core idea of our proposed method is to tell which grids contribute more to the global (dis)similarity with the importance heatmap.
In \cite{TED}, the authors mentioned that there are many challenges to provide AI explanations, such as the lack of one satisfying formal definition for effective human-to-human explanations. However, \cite{LIME} outlines four desirable characteristics for explanation methods, including interpretable, local fidelity, model-agnostic, and global perspective, and our work manages to satisfy these criteria by constructing interpretable maps with local information in the field of face verification.
In \cite{XFR_eccv/WillifordMB20}, the proposed “inpainting game” takes the triplet image pair to investigate how much the saliency map overlaps the ground-truth inpainting mask. It provides a novel metric to evaluate the quality of the attention map in one explainable face recognition system. Our work, however, is to generate new kinds of similarity/attention maps specifically for the face verification problem, not to invent a way to measure the quality of explainability.

 The most related work is \cite{towards-interpretable-face-recognition}. In this work, the authors applied the spatial activation diversity loss and the feature activation diversity loss to learn more structured face representations and force the interpretable representations to be discriminative. Their definition of interpretability of the face representation is that each dimension of the representation can represent a face structure or a face part. Nevertheless, the visualization produced by their method cannot accentuate dominant filters or responses in the face verification task because it is conditioned on a single image instead of one verification pair. Compared to \cite{towards-interpretable-face-recognition}, our model can provide both the quantitative and qualitative reasons that explain why two face images are from the same person or not. If the two face images are viewed as the same person by the model, our proposed method can clearly show which patches on the face are more representative than others via providing local similarity values and the attention weights. 
 

\section{Proposed Approach}

First, we define the ideal properties of $xCos$ metric. Second, we propose three possible $xCos$ formulas.

\subsection{Ideal xCos Metric}
Compared with the traditional cosine similarity for face verification, the ideal $xCos$ (explainable cosine) metric should not only output a single similarity score but also produce \textbf{spatial explanations} on it. That is, $xCos$ should enable humans to understand why the two face images are from the same person (or not) by showing the composition of $xCos$ value in terms of \textbf{components that make sense to humans (e.g., their noses look similar)}. Besides this explainable property, face verification models using $xCos$ as the metric should remain good performance so that it could be used to replace cosine metric in real scenarios.

\subsection{xCos Candidates.}\label{section:xCosCandidates}
Given a face image $I$ and a CNN feature extractor $C$, we can get the grid features $F_{I}$ of size $(h_{F}, w_{F}, c_{F})$:
\begin{equation} \label{eq:2}
    F_{I} = C(I)\in\mathbf{R}^{h_{F}, w_{F}, c_{F}}
\end{equation}
The overall similarity score of $xCos$ is the weighted sum of local similarities, and the weights are from the attention map $\textbf{W}$.
To concisely demonstrate the core idea of $xCos$, we first formulate $xCos$ metric as a general function of $F_{A}$, $F_{B}$, and $\textbf{W}$:
\begin{equation}\label{eq:xCosGeneralFomulation}
    xCos(F_{I_{A}}, F_{I_{B}}, \textbf{W}) = \sum_{i=1}^{h_{F}}\sum_{j=1}^{w_{F}}w_{i,j} *cos(F^{i,j}_{I_{A}},F^{i,j}_{I_{B}})
\end{equation}
where $F_{I}^{i, j}$ is the grid feature at position $(i, j)$, $\textbf{W}\in\mathbf{R}^{h_{F}, w_{F}}$ is the attention matrix, $w_{i, j}\in\textbf{W}$ is the attention weight at position $(i, j)$, and $I_{A}, I_{B}$ refer to two different face images A and B. Three candidates are proposed for the $xCos$ metric:

\subsubsection{Patched $xCos$}\label{patchedxCos}

The most intuitive $xCos$ implementation is to set equal importance for each grid. 
This $xCos$ candidate simply realizes the idea that every pair of the grids on faces should be similar if the two faces are from the same person. 
By comparing the patched $xCos$ with the following $xCos$ variants, we can know whether every grid in the spatial feature shares the same importance.
We let \textbf{unit attention U}:
\begin{equation}
    \textbf{U} = \frac{1}{h_{F}* w_{F}} \textbf{J}_{h_{F}, w_{F}}
\end{equation}
where $\textbf{J}_{h_{F}, w_{F}}$ is the all-ones matrix of size $(h_{F}, w_{F})$, and the patched $xCos$ can be calculated in this way:
\begin{equation}
    xCos_{patched} = xCos(F_{I_{A}}, F_{I_{B}}, \textbf{U})
\end{equation}

\subsubsection{Correlated-patched $xCos$}

Inspired by \cite{FG-18_Castan2018VisualizingAQ}, the facial information is contained majorly around the nose and the periocular region, so there exists an unequal amount of information for different parts of the face. Therefore, we come up with a method to extract the overall importance level for different parts. By calculating the correlation weights of the overall pair similarities and similarities of a given patch, we can get a rough idea of whether the local similarity for certain face parts can represent the global similarity.
We can change the unit attention to correlated-attention \textbf{P}, with the global face features $f_{I_{C}}, f_{I_{D}}$ extracted from any target deep face verification model:
\begin{equation}
    \textbf{P} \in \mathbf{R}^{h_{F}, w_{F}} 
\end{equation}
where the element $p^{i, j}$ in $\textbf{P}$ is the Pearson correlation of the set 
\begin{equation}
\left \{(cos(F^{i, j}_{I_{C}}, F^{i, j}_{I_{D}}), cos(f_{I_{C}}, f_{I_{D}}))  \right\}
\end{equation} over all the image pairs $(I_{C}, I_{D})$ in the training dataset (C, D are arbitrary identity indices in the dataset).
As a result, we get the formula of correlated-patched $xCos$:
\begin{equation}
    xCos_{corr} = xCos(F_{I_{A}}, F_{I_{B}}, \textbf{P})
\end{equation}

\subsubsection{Attention-patched $xCos$}

The attention-patched $xCos$ enables the attention to be conditioned on the input image pair. This design is beneficial when the attention module needs to highlight or de-emphasize some parts of the images. For example, the attention weights for where the mask is put on should be decreased, and the attention weights for salient characteristics like big eyes or tiny mouths should be increased.
Therefore, we propose another kind of $xCos$ metric which learns the attention \textbf{L}, i.e.
\begin{equation}
    \textbf{L} = M_{attention}(F_{I_{A}}, F_{I_{B}}) \in \mathbf{R}^{h_{F}, w_{F}}
\end{equation}, where $M_{attention}$ is a CNN module. The learned attention $\textbf{L}$ is supervised by the cosine similarity of $f_{I_{A}}$ and $f_{I_{B}}$ that are generated with any target face verification model. With this module, we can formulate the attention-patched $xCos$ as follows:
\begin{equation}
    xCos_{attention} = xCos(F_{I_{A}}, F_{I_{B}}, \textbf{L})
\end{equation}
\subsection{Network Architecture}

For current face verification models, the main obstacle to interpretability is that the fully connected layer removes the spatial information, so it is hard for humans to understand how the convolution features before the fully connected layer are combined in a human sense. To address this problem, we propose a two-streamed network with a slightly different backbone and one plug-in $xCos$ module, as described in the following sections:

\subsubsection{Backbone Modification} We attempt to learn the face representation which is not only discriminative but also spatially informative. To achieve this goal, we choose the backbone of the target face recognition model, called $f(C'(I))$, delete its fully-connected part $f(x)$ for face feature extraction, and then append the 1 by 1 convolutional layer $C_{1x1}$ after the original convolutional layers $C'(I)$, i.e. the $C(I)$ in the previous subsection is equal to $C_{1x1}(C'(I))$.
The resulting feature $F_{I}$ plays two roles:
\begin{enumerate}
    \item When it is flattened, $F_{I}$ represents the entire face.
    \item When it is viewed as the grid features, the local information of every grid $F_{I}^{i, j}$ is used to compute local similarities and attention weights.
\end{enumerate}
\subsubsection{Patched Cosine Calculation}
Given a pair of face convolutional features, $F_{I_{A}}, F_{I_{B}}$, each of size $(h_{F}, w_{F}, c_{F})$, the proposed method computes the cosine similarity in each grid pair and generates a patched cosine map $\textbf{S}\in\mathbf{R}^{h_{F}, w_{F}}$. Each element in this map $\textbf{S}$ represents the similarity of each corresponding grid. With this patched cosine map $\textbf{S}$, we can inspect which parts of the face images are considered similar by the model.
\subsubsection{$xCos$ Calculation}
Given two convolutional feature maps, $F_{I_{A}}$, $F_{I_{B}}$, we can first compute the patched cosine map $\textbf{S}$ and generate the attention map $\textbf{W}\in\{\textbf{U}, \textbf{P}, \textbf{L}\}$. Then, we perform the Frobenius inner product $<\textbf{S}, \textbf{W}>_{F}$ to get the value of $xCos$. Specifically, we sum over the results of element-wise multiplication on the attention map $\textbf{W}$ and the patched cosine map $\textbf{S}$, and then obtain the $xCos$ value defined in \ref{section:xCosCandidates}.
\subsubsection{Attention on Patched Cosine Map}
Given two face images, $I_{A}$ and $I_{B}$, we compute their cosine similarity with any target face verification model, i.e. let $c' = cos(f_{I_{A}}, f_{I_{B}})$. 
Then, the attention module $M_{attention}$ can be learned with two feature maps $F_{I_{A}}, F_{I_{B}}$ and the supervising cosine score $c'$.

Inside $M_{attention}(F_{I_{A}}, F_{I_{B}})$, we use convolution layers to perform dimensionality reduction for the two face features $F_{I_{A}}, F_{I_{B}}$, and then fuse the 2 deduced features by the concatenation along the channel dimension. Next, we feed the fused feature into two convolution layers, normalize the output feature map, and get the attention map $\textbf{L}\in\mathbf{R}^{h, w}$.

After getting $\textbf{L}$, we apply element-wise multiplication on the attention map $\textbf{L}$ and the patched cosine map $\textbf{S}$, sum the results to get the $xCos_{attention}$ with value $c$, and calculate the L2-Loss of $c$ and $c'$ so that $\textbf{L}$ is trainable.

\subsubsection{Multitasking for Two-branched Training}
\label{sec:2-branch-training}
As shown in Fig.~\ref{fig:architecture}, the proposed method contains two branches, including the identification branch and the $xCos$ branch. 

The identification branch is trained with the flattened 1 by 1 convolution feature $F_{I_{A}}, F_{I_{B}}$, and the loss function for the identification task, $\mathcal{L}_{id}$, can be the one from ArcFace \cite{deng2018arcface}, CosFace \cite{Wang2018CosFaceLM}, or any target deep face recognition model. Take ArcFace \cite{deng2018arcface}, for example, the $\mathcal{L}_{id}$ is:

\begin{equation}
\mathcal{L}_{id} =  -\frac{1}{N}\sum_{i=1}^{N}log\frac{e^{s(cos(\theta_{y_i} + m))}}{e^{s(cos(\theta_{y_i} + m))} + \sum_{j = 1, j \neq y_i}^{n}e^{s(cos(\theta_{j}))}}
\end{equation}

, where $N$ is the batch size, $y_i$ denotes the $i$-th identity class, $s$ is the normalized embedding feature for the input image, $\theta_{y_i}$ is the angle between the $i$-th class embedding and the input embedding, and $m$ is the angular margin penalty.

The $xCos$ branch performs the task of regressing the $xCos$ value $c$ to the cosine value $c'$ calculated from the target model, and $\mathcal{L}_{cos}$, the loss of regressing $xCos$ to cosine value, is L2-Loss:
\begin{equation}
\mathcal{L}_{cos} =  \frac{1}{N'}\sum_{n=1}^{N'}(c_{n} - c'_{n})^{2}
\end{equation}
, where the $N'$ refers to the number of image pairs in each batch, and $n$ denotes the $n$-th pair in one batch.

The overall loss function for the two-branched training is:

\begin{equation}
\mathcal{L} = \mathcal{L}_{cos} + \lambda \cdot \mathcal{L}_{id}
\end{equation}

, where $\lambda$ is the trade-off weight and $\lambda$ = 1 is chosen in all experiments below.
$\mathcal{L}_{cos}$ guides the regression of $xCos$ value, while $\mathcal{L}_{id}$ makes the identity feature more discriminative.
\section{Experiments}
\subsection{Implementation Details}
\subsubsection{Datasets} We use publicly available MS1M-ArcFace \cite{Guo2016MSCeleb1MAD,deng2018arcface} as training data, and use LFW \cite{LFWTech}, AgeDB-30 \cite{AgeDB} \cite{Deng2017MarginalLF}, CFP \cite{CFP}, CALFW \cite{calfw}, VGG2-FP \cite{cao2018vggface2}, AR database \cite{ARdatabase}, and YTF \cite{YTF_Wolf2011FaceRI} as our testing datasets.

\subsubsection{Data Preprocessing}
We follow the data preprocessing pipeline that is similar to \cite{deng2018arcface, Wang2018CosFaceLM, SphereFace}. We first use MTCNN \cite{MTCNN} to detect faces. Then we apply similarity transform with 5 facial landmark points on each face to get aligned images. Next, we randomly horizontal-flip the face image, resize it into 112 x 112 pixels, and follow the convention \cite{DiscriminativeFeatureLearning, Wang2018CosFaceLM} to normalize each pixel (in [0, 255] for each channel) in the RGB image by subtracting 127.5 then dividing by 128.  

\subsubsection{CNN Setup}
We mainly apply the same backbone as the one in ArcFace \cite{deng2018arcface}. However, we replace the last fully connected layer and the flatten layer before it with the 1 by 1 convolutional layer (input channel size = 512; output channel size = 32), and call the output of it as grid features $F_{I}$. A RGB image $I$ of size (112, 112, 3) will result in a grid feature $F_{I}$ of size (7, 7, 32). When training the face identification branch, we flatten the grid feature $F_{I}$ into a 1-D vector with dimension 1568.

\subsubsection{$xCos$ Module Setup}
Given two grid features, $F_{I_{A}}, F_{I_{B}}$,  of size (7, 7, 32), our goal is to produce one attention map $\textbf{L}$ and one patched cosine map $\textbf{S}$. The attention map $\textbf{L}$ is obtained by performing convolution over the fused grid features. First, we use a convolution layer with kernel size = 3 and padding = 1 to perform dimension reduction on $F_{I}$ with the output channel dimension = 16. The two reduced convolution features of size (7, 7, 16) are then concatenated into a new fused grid feature of size (7, 7, 32). Second, we feed the fused grid feature into another two convolution layers to get the output $\textbf{L}$, of size (7, 7). Finally, we normalize the attention map with a softmax function to make sure the sum of all the 49 grid attention weights is 1. The patch-cosine map $\textbf{S}\in\mathbf{R}^{7, 7}$ is obtained by computing the grid-wise cosine similarity between any paired grid features from $F_{A}$ and $F_{B}$. The $xCos$ value is calculated by performing the Frobenius inner product between $\textbf{L}$ and $\textbf{S}$. The learning rate is 1e-3 for all models, and it is divided by 10 after 12, 15, 18 epochs.
\subsection{Quantitative Results}
\subsubsection{Face Verification Performance}
\begin{table}
    \centering
    \begin{tabular}{cc}
        Method & Accuracy \\
        \hline\hline
        Human performance \cite{Human_LFW_performance} & 97.53\% \\
        GaussianFace \cite{GaussianFace} (non-Deep) & 97.79\% \\
        CosFace \cite{Wang2018CosFaceLM} & 99.33\% \\
        ArcFace \cite{deng2018arcface}& 99.83\% \\
        \hline
        \textbf{attention-patched $xCos$} (Ours, CosFace) & 99.67 \% \\
        \textbf{attention-patched $xCos$} (Ours, ArcFace) & 99.35 \% \\
    \end{tabular}

    \caption{\textbf{Face verification accuracy on LFW dataset.} Compared to other face verification models, the proposed $xCos$ module significantly improves explainability with a minimal drop of performances.}
    \label{tab:LFW_accuracy}
\end{table}
To demonstrate the effectiveness of our proposed method, we show the performance of $xCos$ in the Table~\ref{tab:LFW_accuracy}. From Table~\ref{tab:LFW_accuracy}, we can observe that the $xCos$ module not only provides explainability with the trade-off of a little drop of accuracy but also produces promising performance gain over the human performance and some earlier non-deep face verification models like GaussianFace \cite{GaussianFace}.

\begin{table*}
    \centering
    \begin{tabular}{ |l|c|c|c|c|c|c|c|c| } 
    \hline
     BackBone& \multicolumn{4}{|c|}{ArcFace \cite{deng2018arcface}}& \multicolumn{4}{|c|}{CosFace \cite{Wang2018CosFaceLM}} \\ \hline
     Methods & baseline* & patch. & corr.   & atten.  & baseline* & patch.  & corr.   & atten. \\
     \hline
     Feature Layer & FC & \multicolumn{3}{|c|}{1x1} & FC & \multicolumn{3}{|c|}{1x1} \\
     \hline
     Attention Type & - & $\textbf{U}$ &   $\textbf{P}$ & $\textbf{L}$ & - & $\textbf{U}$ & $\textbf{P}$ & $\textbf{L}$ \\
     \hline
     LFW \cite{LFWTech} (\%) & \textbf{99.45}& 99.23 & 99.12 & 99.35 & 99.28 & 99.63& 99.60 & \textbf{99.67}\\
     YTF \cite{YTF_Wolf2011FaceRI} (\%)& 95.06 & 95.50 & \textbf{95.56} & 95.50 & 96.24 & 96.92 & 96.92 & \textbf{96.92}\\
     VGG2-FP \cite{cao2018vggface2} (\%)& 89.94 & 91.14 & \textbf{91.22} & 90.54 & 91.86 & \textbf{93.66} & 93.66 & 93.38\\
     AgeDB-30 \cite{AgeDB} \cite{Deng2017MarginalLF} (\%) & 91.60 & 92.47 & 92.73 & \textbf{93.81} & 89.60& 95.20 & 95.28 & \textbf{95.93}\\
     CALFW \cite{calfw} (\%)& 92.55& 93.23 & 93.17 & \textbf{94.08} & 91.30 & 94.83 & 94.77 & \textbf{95.10}\\
     CFP-FF \cite{CFP} (\%)& 99.08& 99.09 & 99.13 & \textbf{99.31} & 98.80 & 99.44 & 99.44 & \textbf{99.44}\\
     CFP-FP \cite{CFP} (\%)& 87.56 &88.60 & \textbf{88.64} & 88.08 & 90.61 & 93.07& 93.16 & \textbf{93.54}\\
     \hline
    \end{tabular}
    \caption{\textbf{Ablation Studies.} The patch., corr., and atten. refer to the patched $xCos$, correlated-patched $xCos$, and attention-patched $xCos$ mentioned in Section~\ref{section:xCosCandidates}, respectively; ArcFace \cite{deng2018arcface} and CosFace \cite{Wang2018CosFaceLM} represent common backbone models used in face identification. From this table, we can observe that (1) $xCos$ brings explainability without degrading the performance; (2) The plug-in $xCos$ attention module can perform well in different face verification backbones. Note (*): We train the baseline with the same training setting for $xCos$ and turn off the testing time augmentation to have a fair comparison.}
    \label{tab:Ablation}
\end{table*}

\subsubsection{Ablation Studies}
As shown in Table~\ref{tab:Ablation}, we use the face recognition models without the backbone modification as baseline, and then observe the effectiveness of $xCos$ via applying different attention weights $\textbf{W}\in\{\textbf{U}, \textbf{P}, \textbf{L}\}$. In pursue of a fair comparison, we train the baseline with the same setting of $xCos$ except the feature extraction layer, and turn off the testing time augmentation for the baseline because it will apply an averaging operation over features, which leads to the mix of spatial information for our convolutional features. 
Among most of the testing datasets, attention-patched $xCos$ achieves the best performances, suggesting that our attention module takes effect. However, in datasets VGG2-FP \cite{cao2018vggface2} and CFP-FP \cite{CFP}, it seems that the patched $xCos$ and the correlated-patched $xCos$ may get a better result than the attention-patched $xCos$. We hypothesize that our proposed models, which are trained on aligned face images, do not perform as expected due to the huge pose difference and pose variations in these two datasets. 
Also, both the baseline model and our purposed $xCos$ models have noticeable performance drops between the pose-varying datasets and datasets without pose variations. Therefore, we believe this is a general issue for all the face verification models which do not handle pose variations by design. We discuss how to optimize both the explainability and the model performance in Section~\ref{Frontalization}.

\subsubsection{Computational Cost}
Although there are some additional costs to calculate the pairwise cosine similarity and attention map in our system, the feature extraction process is still the computational bottleneck. When ignoring all disk reading and writing time and running on an i7-3770 CPU with a 1080ti GPU, the inference for a pair of faces takes 6.1 ms and 6.7 ms for the original model and our $xCos$ model, respectively. Compared to the explainability gain over the original model, this efficiency drop is negligible.

\subsubsection{The effectiveness of regressing xCos to the cosine value}
\begin{figure}
\includegraphics[width=.5\linewidth]{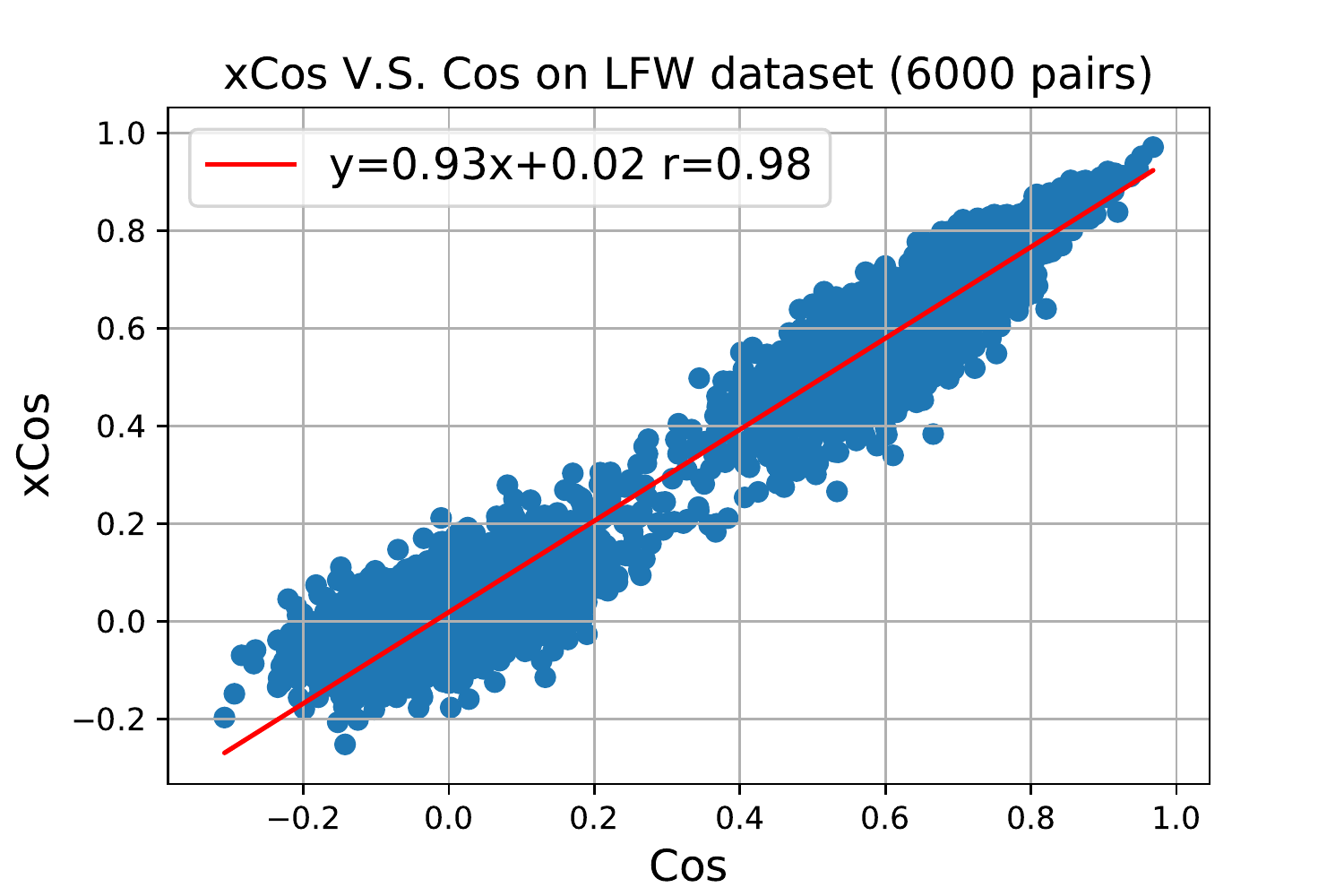}
\caption{\textbf{The correlation between the xCos and the cosine value.} For each pair of photos in the LFW \cite{LFWTech} dataset, the xCos value and the cosine similarity are computed from the proposed model and the pretrained ArcFace \cite{deng2018arcface} model respectively. The high correlation coefficient (r=0.98) shows that the $xCos$ branch of the proposed model learns from the existent ArcFace model.}
\label{Fig:corr-btw-xCos-and-cos}
\end{figure}

In Section~\ref{sec:2-branch-training}, the output of $xCos$ branch is regressed with the cosine value of the target face verification model. Fig.~\ref{Fig:corr-btw-xCos-and-cos} demonstrates the effectiveness of the regression task on the LFW \cite{LFWTech} dataset. By correlating the similarity scores, the spatial maps generated from the $xCos$ branch can be one interpretation of how the target model produces the verification result.

\subsection{Qualitative Results}
\begin{figure*}[!ht]
  \includegraphics[width=\textwidth]{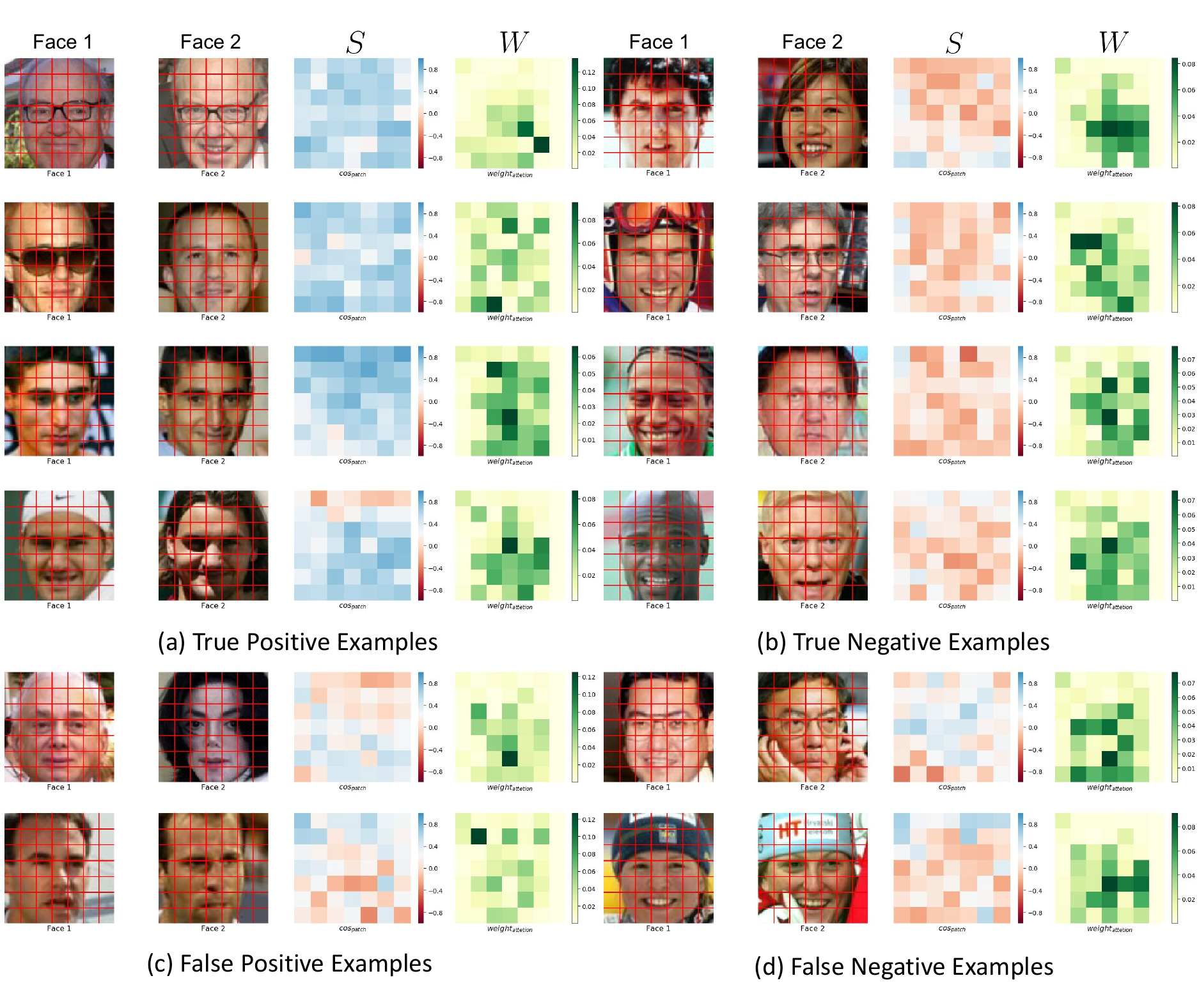}
  \caption{\textbf{Qualitative Results.} The third and the fourth columns of each example represent the patched cosine similarity maps and the attention weights maps. In the fourth row of (a), our model pays attention (green grids in the $\textbf{W}$) to the similar shapes of the two noses (blue grids in the $\textbf{S}$), rather than the different hairstyles (red grids in the $\textbf{S}$). In the first row of (d), it is clear that the hands distracts the model. With the visualizations, we can alarm users to put their hands away to avoid verification failure. With the aid of our proposed cosine similarity map $\textbf{S}$ and attention map $\textbf{W}$, we can easily interpret the visualized results in the confusion matrix. Thus, users can be more confident to know when models go right (or wrong), and $xCos$ can play a role in helping optimize the design of the face verification model.}
  \label{fig:Qualitative}
\end{figure*}
\begin{figure*}[!ht]
    \includegraphics[width=1.\linewidth]{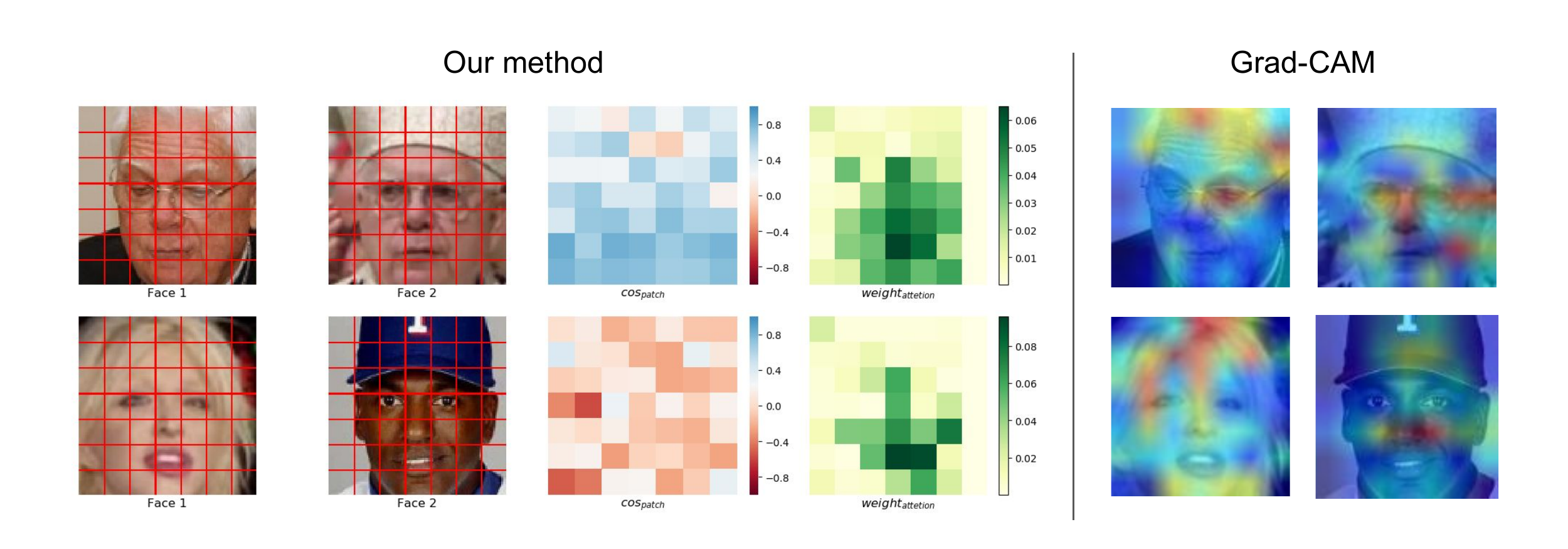}
    \caption{\textbf{Comparison with Saliency Methods} (1) The first row shows one true positive pair. It is interpretable with the proposed $xCos$ that the forehead area is not similar and not important for the verification result, while it is hard for a human to interpret how the two individual heat maps around the forehead contribute to the result by applying saliency methods like Grad-CAM \cite{Grad-CAM} on the ArcFace \cite{deng2018arcface} model. (2) The second row is one true negative pair. The saliency method just puts the most significant pixels side by side, while our method reveals that the dissimilarity caused by the cap is not important for the $xCos$ model. Both pairs are from the LFW \cite{LFWTech} dataset. }
    \label{Fig:grad-CAM}
    
\end{figure*}
\subsubsection{Visualizations of $xCos$}

As shown in Fig.~\ref{fig:Qualitative}, there are two interesting phenomena worth mentioning:
\begin{enumerate}
    \item The area around central columns is of great interest to the $xCos$ model. By observing the weight distributions on the attention maps, we can conclude that the central convolution feature is influential for the model to verify the identity.
    \item The area near mouths and chins is of greater importance than the upper parts of faces. People may wear hats, change hairstyles, or become bald as growing older, so the model pays less attention to the area on the top of faces. On the contrary, the variations of the shape of mouths and chins are constrained to the color of lips or facial expression like smiling. For instance, the fourth row in Fig.~\ref{fig:Qualitative}(a) and the second row in Fig.~\ref{fig:Qualitative}(d) both contain faces with hats, while the model pays less attention to those facial parts which are occupied with hats. 
\end{enumerate}
\subsubsection{Comparison with saliency methods.}

\begin{table}
    \begin{tabular}{|c|c|c|c|}
    \hline
    & \begin{tabular}[c]{@{}c@{}}local \\ importance\end{tabular} & \begin{tabular}[c]{@{}c@{}}local \\ similarity\end{tabular} & \begin{tabular}[c]{@{}c@{}}verification \\ metric\end{tabular}   \\ \hline
    $xCos$& V & V & $xCos$ + $\textbf{S}$ + $\textbf{W}$\\ \hline
    saliency maps* & V& X  & cosine value\\ \hline
    
    \end{tabular}
    \caption{\textbf{Differences between $xCos$ and saliency maps.} $\textbf{S}$ and $\textbf{W}$ are the interpretable maps defined in the paper. Note (*): saliency maps are methods whose outputs are two individual heat maps for one verification pair.}
    \label{tab:table_differences}
\end{table}
Saliency methods like Grad-CAM \cite{Grad-CAM} provide attention-like heat maps. However, it is mainly for identification tasks but not verification tasks. Fig.\ref{Fig:grad-CAM} shows four qualitative results of Grad-CAM. It is hard for us to interpret why the two face images are verified as the same person or not.
Several previous works have dealt with finding the pixels that contribute the most. However, those works, even the most relevant one \cite{towards-interpretable-face-recognition}, (1) provide no \textbf{local similarity} information in their saliency maps and (2) hardly focus on the face verification task. (See Table~\ref{tab:table_differences}.) 
Contrarily, $xCos$ not only highlights essential regions but tells users which grids are (dis)similar. Revealing local similarity helps users debug the verification system, for example, by showing the local dissimilarities caused by hand occlusion (e.g., the first row of Fig~\ref{fig:Qualitative}(d)).

\section{Discussions}
\subsection{Additional Robustness to Occlusion} 
\begin{figure*}[!ht]
\includegraphics[width=1.\linewidth]{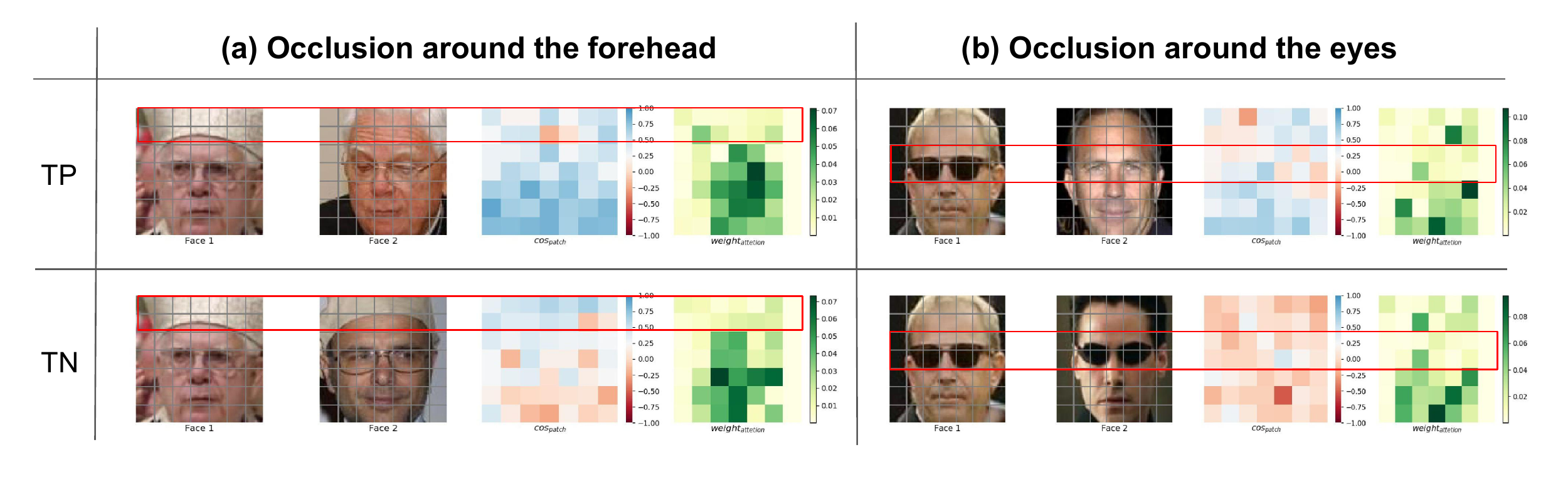}

\caption{\textbf{The Patched Cosine Maps and the Attention Matrices of two Occluded Face Triplets.} The “TP”(True Positive)/“TN”(True Negative)  row compares one image to one positive/negative image, respectively. The visualizations for the triplet images reveal which grid (dis)similarities are important. In (a), the attention matrices show that the occlusion around the forehead is not important, but the (dis)similarities around the nose, chin, and eyes are essential for the verification. In (b), the dissimilarities around the eyes do not affect the verification score a lot. With the xCos module, the verification result can be interpreted with local similarities and attention weights.
}
\label{Fig:triplet-visulization}
\end{figure*}
\begin{figure*}[!ht]
\includegraphics[width=1.\linewidth]{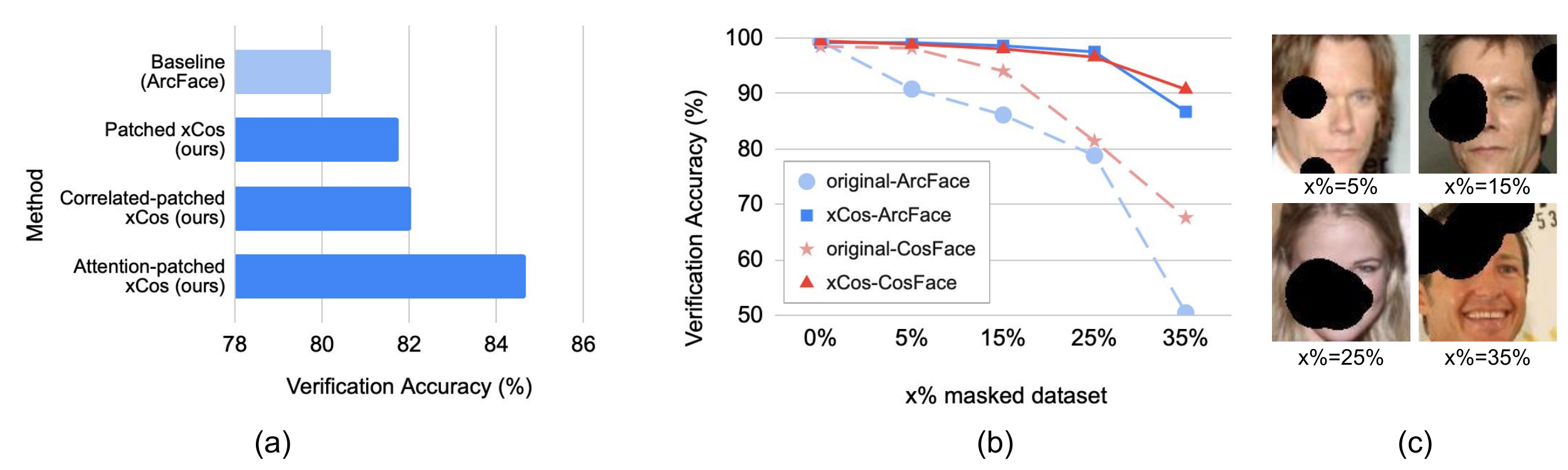}
\caption{\textbf{(a) Face Verification Accuracy on the Occlusion Subset of AR Database \cite{ARdatabase}.} The proposed $xCos$ method provides not only explainability but also additional robustness to partially occluded faces. \textbf{(b) Face Verification Accuracy on the x\% Masked LFW \cite{LFWTech}} Free-form masks \cite{chang2019free-form} are applied on the images of LFW dataset. \textbf{(c) Examples of the x\% Synthetic Occlusion Dataset.} The proposed $xCos$ has less performance drop than common face recognition models, including ArcFace and CosFace.}
\label{Fig:occlusion-AR-and-free-form}
\end{figure*}

Since the local similarities are independently calculated and the learned attention is conditioned on the input image pair, our method should be more robust than the original model when faces are partially occluded. For instance, the occlusions around the forehead and the eyes hardly contribute to the verification result in Fig.~\ref{Fig:triplet-visulization}. We quantitatively test the robustness to occlusion on two datasets. AR face \cite{ARdatabase} is a natural occlusion face database with around 4K faces of 126 subjects and thus it is a good test set for the occlusion experiment. We select the faces with scarfs or glasses and exclude those which can not be detected by MTCNN \cite{MTCNN}. After the selection, 1488 images are used to randomly generate 6000 positive pairs and 6000 negative pairs. As shown in Fig.~\ref{Fig:occlusion-AR-and-free-form}(a), our proposed methods outperform the original ArcFace model even without the attention module. Besides, we use the free-form masks in \cite{chang2019free-form} to create synthetic CASIA \cite{CASIA-Yi2014LearningFR} and LFW \cite{LFWTech} occlusion datasets for fine-tuning and testing, respectively. There is one mask that occupies about x\% out of the total area for each image in the training or testing dataset (See Fig.~\ref{Fig:occlusion-AR-and-free-form}(c) for examples.) From Fig.~\ref{Fig:occlusion-AR-and-free-form}(b), it can be concluded that the proposed $xCos$ method has less performance drop than the original face verification model.

\subsection{How to Evaluate the Quality of the Attention Matrices}
What kind of attention matrix is good remains an open question due to the lack of a universal definition of “good” attention quality. For the synthetic dataset in \cite{XFR_eccv/WillifordMB20}, the \textbf{absolute} quality of the attention matrices can be calculated using the protocol in \cite{XFR_eccv/WillifordMB20}. For a real-world dataset, it is not easy to explicitly evaluate the quality of the attention matrix for a specific verification pair, because there are no human-annotated ground truths for the importance of local similarities. However, there may exist some methods that measure the \textbf{relative} quality of attention matrices. For example, we can determine the relative quality of two types of attention, e.g., the correlated-attention and the learned attention, by comparing the performance of models with them. This measurement is based on the assumption that higher verification performance results from better attention matrices. Since the primary purpose of our proposed method is to design interpretable maps specifically for the face verification problem, we leave the investigation of measuring the absolute quality of attention matrices to future works.

\subsection{How to Adapt xCos From Frontal Images to Profile Ones}\label{Frontalization}
In this work, we open a new avenue for the explainability in the face recognition task. As the pilot study for the emerging problem, we have to take two steps to make our research more convincing: (1) verify that plugging the proposed explainable module into SoTA face recognition models does not degrade the overall verification performance on the ideal test setting (e.g., test on the aligned LFW dataset); (2) Extend the usage of $xCos$ to other rigorous experiment settings, like face images with significant pose variations or extreme illuminations. We are optimistic to see that our work, which realizes the main idea in stage (1), is going to inspire more future research on face applications with critical conditions. 

There are plenty of papers embarked on tackling various challenging conditions, including low light/ resolution settings or large pose variations, cross-age, etc. Following our successful attempt in the first stage, we believe the research communities can adapt the $xCos$ module for many other face recognition problems. For example, some previous works have explored the possibility of recovering the canonical view of face images from non-frontal images using SAE \cite{SPAE_Kan2014StackedPA}, CNN \cite{CNN_FRONTALIZATION}, and GAN \cite{TP_GAN} models, and we can extend the usage of $xCos$ to the cross-pose scenario by performing these preprocessing method first.

\section{Conclusions}
We propose a novel metric for the face verification task, called $xCos$ (explainable cosine). The proposed metric decomposes the overall similarity of two face images into patched cosines and one attention map. With this metric, humans can intuitively understand which parts of the faces are similar and how important each grid feature is. We believe that $xCos$ can be used to inspect the behavior of the face verification model and bridge the gap between the model complexity and human understanding in an explainable way.

\section{Acknowledgments}
This work was supported in part by the Ministry of Science and Technology, Taiwan, under Grant MOST 110-2634-F-002-026 and Qualcomm Technologies, Inc. We benefit from NVIDIA DGX-1 AI Supercomputer and are grateful to the National Center for High-performance Computing.







\bibliographystyle{ACM-Reference-Format}
\bibliography{egbib}










\end{document}